\documentclass[letterpaper, 10 pt, conference]{ieeeconf}  

\IEEEoverridecommandlockouts                              

\usepackage{amsmath,amsfonts}
\usepackage{algorithm}
\usepackage{array}
\usepackage[caption=false,font=normalsize,labelfont=sf,textfont=sf]{subfig}
\usepackage{textcomp}
\usepackage{stfloats}
\usepackage{url}
\usepackage{verbatim}
\usepackage{graphicx}
\usepackage{cite}
\usepackage{color}
\usepackage{amsmath}
\usepackage{amssymb}
\usepackage{booktabs}
\usepackage{multirow}
\usepackage{mmstyle}
\usepackage[noend]{algpseudocode}

\title{\LARGE \bf
Enhancing Human-Centered Dynamic Scene Understanding via Multiple LLMs Collaborated Reasoning
}

\author{Hang Zhang$^{1}$, Wenxiao Zhang$^{1}$, Haoxuan Qu$^{1}$, and Jun Liu$^{1*}$
\thanks{* Corresponding Author}
\thanks{$^{1}$Hang Zhang, Wenxiao Zhang, Haoxuan Qu, and Jun Liu are with the Information Systems Technology and Design (ISTD) pillar, 
        Singapore University of Technology and Design (SUTD), 8 Somapah Road, 487372, Singapore. 
        {\tt\small \{hang\_zhang, jun\_liu\}@sutd.edu.sg, haoxuan\_qu@mymail.sutd.edu.sg, wenxxiao.zhang@gmail.com}}%
}

\begin{document}

\maketitle
\thispagestyle{empty}
\pagestyle{empty}

\begin{abstract}
Human-centered dynamic scene understanding plays a pivotal role in enhancing the capability of robotic and autonomous systems, in which Video-based Human-Object Interaction (V-HOI) detection is a crucial task in semantic scene understanding, aimed at comprehensively understanding HOI relationships within a video to benefit the behavioral decisions of mobile robots and autonomous driving systems. Although previous V-HOI detection models have made significant strides in accurate detection on specific datasets, they still lack the general reasoning ability like human beings to effectively induce HOI relationships. In this study, we propose V-HOI Multi-LLMs Collaborated Reasoning (V-HOI MLCR), a novel framework consisting of a series of plug-and-play modules that could facilitate the performance of current V-HOI detection models by leveraging the strong reasoning ability of different off-the-shelf pre-trained large language models (LLMs). We design a two-stage collaboration system of different LLMs for the V-HOI task. Specifically, in the first stage, we design a Cross-Agents Reasoning scheme to leverage the LLM conduct reasoning from different aspects. In the second stage, we perform Multi-LLMs Debate to get the final reasoning answer based on the different knowledge in different LLMs. Additionally, we devise an auxiliary training strategy that utilizes CLIP, a large vision-language model to enhance the base V-HOI models' discriminative ability to better cooperate with LLMs. We validate the superiority of our design by demonstrating its effectiveness in improving the prediction accuracy of the base V-HOI model via reasoning from multiple perspectives. 
\end{abstract}

\begin{figure}
    \centering
    \includegraphics[width=\linewidth]{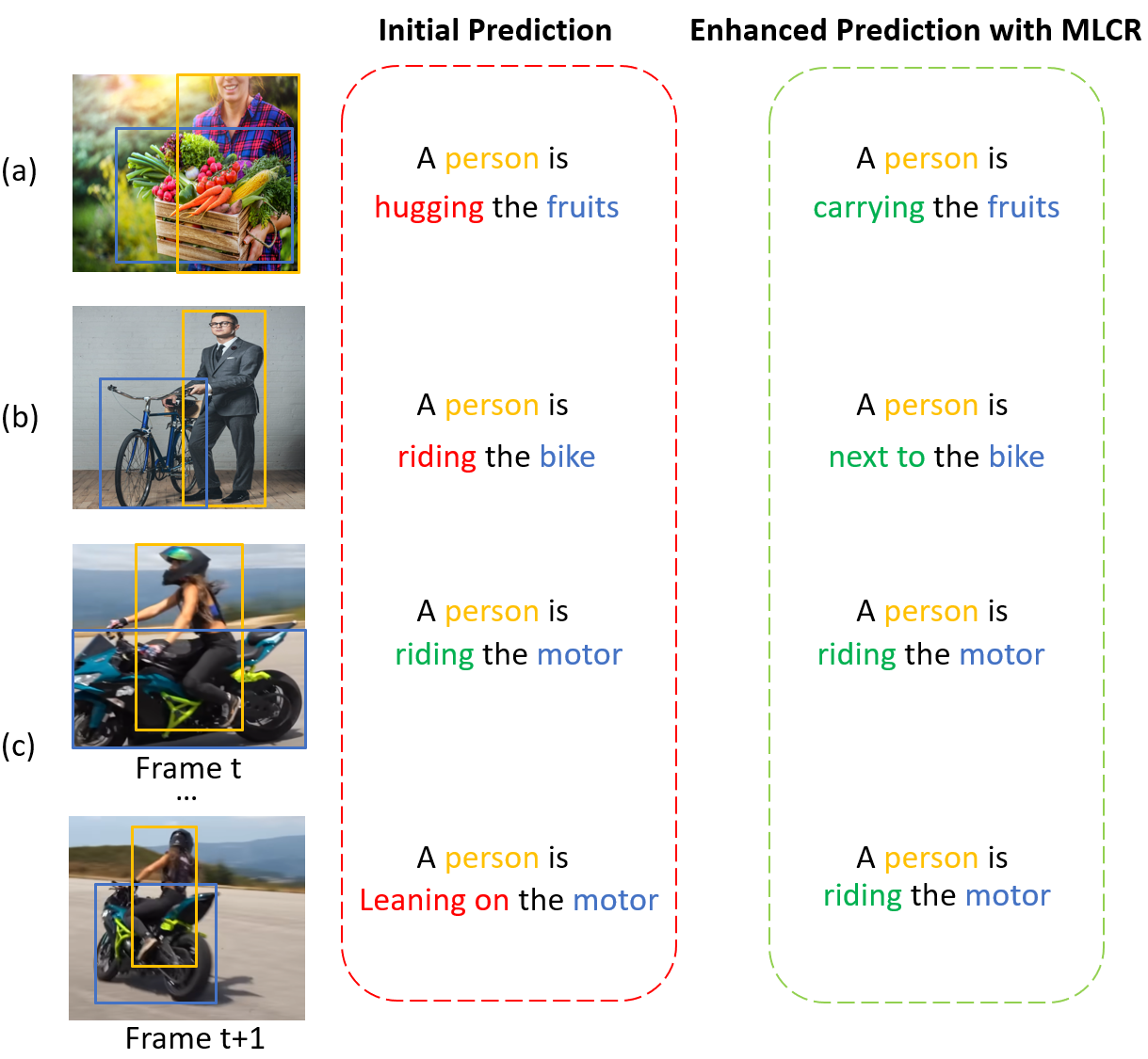}
    \caption{The initial prediction from the SOTA V-HOI model\cite{ni2023human} will cause incorrect relation prediction. Instead, our proposed MLCR refined the prediction to get the correct results.}
    \label{fig:intro}
\end{figure}

\section{Introduction} 
Dynamic scene understanding plays a crucial role in ensuring safe and reasoned behavior planning and decision-making for robots and autonomous vehicles \cite{tr2, li2022embodied, chang2021comprehensive}. It enables these intelligent systems to interpret and respond to Human-Object Interactions (HOI) in dynamic environments effectively \cite{li2023knowledgedriven}. Accurate video-based HOI detection in human-centered dynamic scene understanding stands as a fundamental milestone in advancing our comprehension of complex scenes, involving the localization of human-object pairs and the recognition of interaction labels \cite{ni2023human}. 

Earlier research predominantly focused on HOI detection within static images. For example, \cite{kim2020detecting, Li_2020_CVPR, li2020hoi, ulutan2020vsgnet, xu2022meta,zhong2020polysemy} typically follow a two-step methodology. Initially, an object detector is employed to identify the positions of humans and objects. Subsequently, a multi-stream classifier is utilized to predict interactions for each human-object pair. To enhance model efficiency, various end-to-end techniques \cite{liao2020ppdm, wang2020learning, kim2020uniondet, fang2021dirv} have been proposed to simultaneously generate object detection and interaction classes. These image-based methods usually overlook temporal dynamics and lack the ability to identify time-sensitive interactions. To this end, some works \cite{jain2016structural, sunkesula2020lighten, wang2021spatio, sun2021spatial, chiou2021st, ji2021detecting, cong2021spatial, tu2022video} elaborating on Video-based Human-Object Interaction (V-HOI) are proposed to leverage the temporal dependencies between frames and demonstrate superior performance to the image-based methods. 

Despite the great progress yielded for HOI relationship detection, most existing HOI detection models are unable to develop strong common-sense reasoning capabilities compared to human intelligence. Taking the example in Figure \ref{fig:intro} (a), the current SOTA V-HOI model gives the prediction "a person is hugging the fruits", but this does not obey the human common sense where "a person is carrying the fruits" is more suitable. In Figure \ref{fig:intro} (b), the model prediction is "a person is riding the bike." However, the action of "riding" typically occurs when the person is on top of the bike. The example in the picture does not conform to this spatial relationship. In Figure \ref{fig:intro} (c), we show several consistent frames in a video, the initial model prediction is "a person is riding a bike", but in the following frame, the prediction is "a person is leaning on the bike", which is not rational when considering the fact consistency. From here, we could observe that, as trained only on specific datasets, current V-HOI models can have limited common sense reasoning ability, but it could lead to the interaction prediction being performed in better quality when equipped with our method, which aims at enhancing the model reasoning ability (right column). Recently, large language models (LLMs), which contain rich common sense knowledge \cite{guo2023viewrefer, zhao2024large}, have achieved remarkable progress in solving various tasks due to strong reasoning abilities. Also, as different large models are trained on different training corpus, they contain rich yet diverse knowledge. Inspired by the awesome reasoning ability of LLMs, we aim to investigate leveraging the strong reasoning ability of various LLMs to help the V-HOI task. In this work, we propose V-HOI \textbf{M}ulti-\textbf{L}LMs \textbf{C}ollaborated \textbf{R}easoning (\textbf{V-HOI MLCR}), a framework leveraging the rich and diverse knowledge from various LLMs to perform reasoning through a designed collaboration system.

Specifically, we construct the information of spatial layouts and temporal cues as structured instructions, which could enhance LLMs’ ability to "imagine" object locations and the changes in the timeline from merely language prompts. Our design not only enables stable and consistent output structures but also strengthens LLMs’ understanding of the visual concepts of the spatial and temporal attribute value. Moreover, to effectively coordinate this diverse yet beneficial knowledge from different LLMs, we introduce a novel collaboration mechanism among Language Model (LLM) experts. Treating each LLM as a reasoning expert, we implement a two-stage collaboration scheme. In the first stage, our Cross-Agents Reasoning scheme assigns distinct roles to different agents within an LLM, focusing on spatial rationality and temporal consistency reasoning. In the second stage, a cyclic debate mechanism is employed to assess and aggregate responses from various LLMs, enabling the deduction of final answers based on comprehensive knowledge from all LLMs. Additionally, besides leveraging the external knowledge from multiple LLMs to improve the model reasoning ability, we also propose an auxiliary training strategy that utilizes CLIP \cite{radford2021learning} as an internal supervisor to enhance the base V-HOI models' reasoning and discriminative ability in handling ambiguous semantic relations.

In summary, our contributions are as follows:
(I) To the best of our knowledge, we explore leveraging various LLMs to help with V-HOI tasks for the first time. 
(II) We introduce V-HOI MLCR, a novel plug-and-play framework that can facilitate the reasoning abilities of current V-HOI models, through collaborating with multiple large language models as external experts with their rich and diverse knowledge. 
(III) We perform extensive experiments to assess the effectiveness of V-HOI MLCR on two V-HOI detection datasets.

\begin{figure*}
    \centering
    \includegraphics[width=\textwidth]{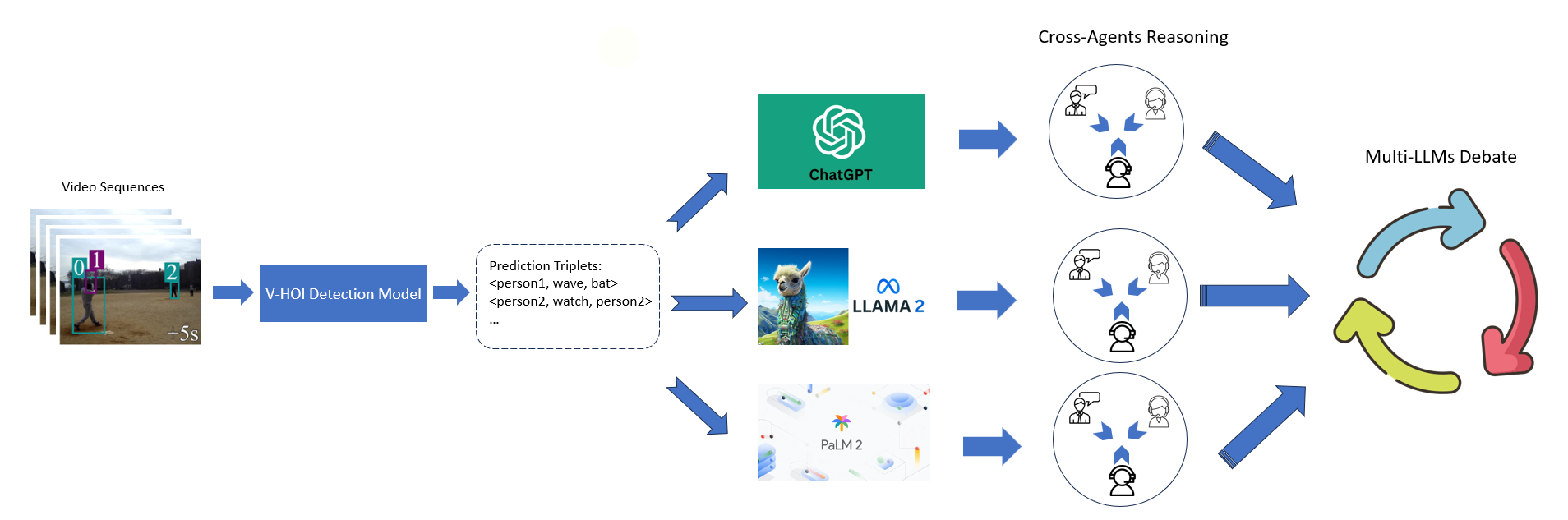}
    \caption{Method Overview. Upon the analysis of video sequences, we initially apply the state-of-the-art models for V-HOI detection to obtain preliminary prediction triplets for individual frames. These triplets are then converted into textual form and processed through our LLMs collaborative framework. Our framework operates in two primary stages: the first stage is Cross-Agents Reasoning, where various distinct agents are established within each different LLM (ChatGPT, LLaMA2, and PaLM2) to evaluate the logic of the predictions from different perspectives, including spatial and temporal coherence. The second stage is the Multi-LLMs Debate, where we integrate responses from various LLMs in a debate-style format to refine and finalize the predictions.}
    \label{fig:overview}
\end{figure*}

\section{Related Works}

\subsection{HOI Detection}

HOI detection is with the challenges of object detection, human-object pairing, and interaction recognition \cite{ning2023hoiclip}. Historically, existing HOI detectors generally fall into either one-stage or two-stage paradigms. The two-stage \cite{kim2020detecting, Li_2020_CVPR, li2020hoi, ulutan2020vsgnet,xu2022meta,zhong2020polysemy} method, which employs a pre-existing detector to identify object locations and classes, succeeded by custom modules for human-object association and interaction recognition. In contrast, the one-stage approach detects HOI triplets by allocating human and object proposals to predefined anchors and then inferring potential interactions \cite{liao2020ppdm, wang2020learning, kim2020uniondet, fang2021dirv}. 

A comprehensive understanding of the dynamic changes in the relationship between a person and an object over a period is pivotal for accurate interaction detection in video analysis \cite{ni2023human}. \cite{jain2016structural} utilizes a Structural Recurrent Neural Network (S-RNN) to depict human-object relations as a spatio-temporal graph and reason interaction sorts. Building upon this, \cite{truong2017structured} enhances the S-RNN by considering object-to-object relationships, while \cite{sunkesula2020lighten} elevates model performance by utilizing learned visual features as graph nodes. In contrast to RNNs, \cite{qi2018learning} introduces the Graph Parsing Network (GPN) for parsing human-object spatio-temporal graphs. Subsequently, \cite{wang2021spatio} devises a two-stream GPN that integrates semantic features. Differing from graph-centric approaches, \cite{sun2021spatial} proposes an instance-based framework for individual human-object pair reasoning, leveraging human skeletons as an added signal for HOIs. Additionally, \cite{chiou2021st} utilizes human pose features for HOI detection, while ST-HOI employs a 3D backbone for precise instance feature extraction from videos. Recently, inspired by the transformer model's success, various instance-based spatio-temporal transformers \cite{ji2021detecting, cong2021spatial, tu2022video} have emerged.

\subsection{Reasoning from Large Language Models}
The process of reasoning stands as a foundational element within human intelligence, pivotal in endeavors like problem-solving, decision-making, and critical thinking. Over recent years, the emergence of large language models (LLMs) has witnessed significant strides in natural language processing. \cite{huang-chang-2023-towards} offers a comprehensive survey delving into the reasoning capacities of LLMs. Within the literature \cite{huang-chang-2023-towards}, the term "reasoning" in the context of LLMs often alludes to a less structured method reliant on intuition, experience, and common sense for drawing conclusions and tackling problems—a skill commonly applied in everyday life. Incorporating grounding information into these expansive language models, as highlighted in \cite{li2023visionfree}, frequently amplifies their reasoning capabilities, proving beneficial across a spectrum of applications.  \cite{ghanimifard2019neural} demonstrates the language model's ability to discern functional and geometric biases of spatial relations through encoding, despite lacking direct access to visual scene features. LayoutGPT \cite{feng2023layoutgpt} utilizes language models to produce spatial arrangements, thereby demonstrating the capabilities of Large Language Models (LLMs) in visual planning. Although previous V-HOI detection models have achieved notable improvements in improving detection accuracy on certain datasets, they still fall short of human-like reasoning abilities to effectively induce human-object interaction relationships, hampered by the biased knowledge obtained from specific training datasets. Hence, LLMs can aid with this through their reasoning capability.

\section{Problem Definition }

Analogous to the HOI detection task in image data \cite{gkioxari2018detecting}, the V-HOI detection task is first proposed in \cite{chiou2021st}, characterized by the retrieval of bounding boxes for the human subjects ${\mathbf{b}_{t, i}}$ and objects ${\mathbf{b}_{t, j}}$, the predicted object classes $c_{t, j}^{obj}$, and the recognition of interaction classes $\mathbf{c}_{t,\langle i, j\rangle}^{inter}$ within each frame $I_t$. Here, $I_t \in \mathbb{R}^{h \times w \times 3}$ represents an RGB frame at time $t$, while the subscripts $i$ and $j$ are used to denote different humans and objects in a frame. The detected HOIs are expressed as a collection of triplets: ${\left\langle\mathbf{b}_{t, i}, \mathbf{c}_{t,\langle i, j\rangle}^{inter}, \mathbf{b}_{t, j}\right\rangle}$, indicating the human (subject), interaction relation class, and object, respectively. To obtain an interaction predicate $\mathbf{c}^{inter}$, current methods usually first use an encoder to get the interaction embedding $f^{inter}$, and then feed it into a prediction head, which is usually a linear layer followed by a softmax function, predicting the probability score of interaction classes ${\mathbf{s}}^{inter} \in [0,1]^{N}$, where $N$ is the number of interaction classes.

\begin{figure*}
    \centering
    \includegraphics[width=\textwidth]{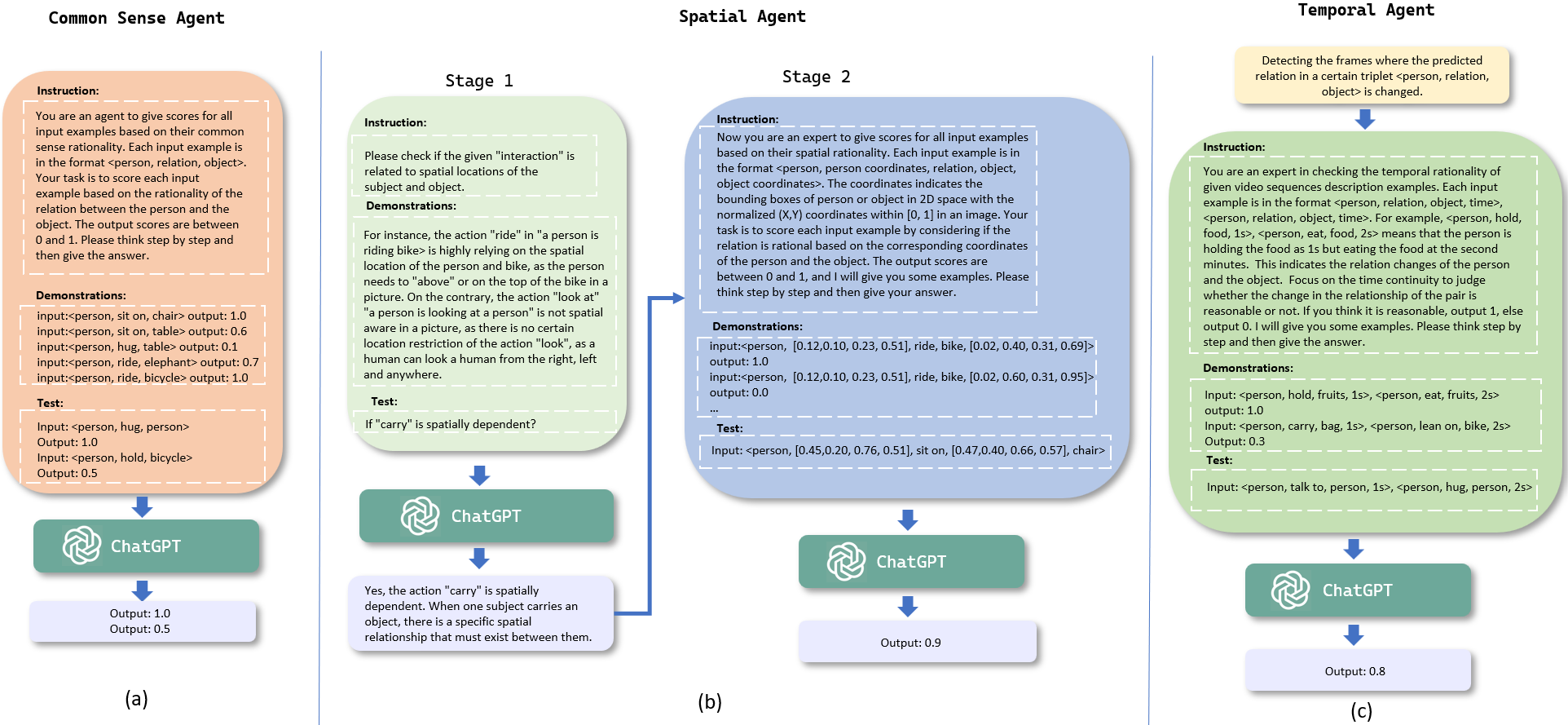}
    \caption {We employ each LLM (e.g. ChatGPT) as a cross-aspect reasoning agent to enhance the accuracy of predictions derived from current V-HOI detection models. Within this framework, we have architected three specialized reasoning agents—those that apply (a)common sense, (b)spatial reasoning, and (c)temporal reasoning to refine the predictions yielded by the extant V-HOI detection model.}
    \label{fig:agents}
\end{figure*}

\section{Method Overview}

Our proposed V-HOI Multi-LLMs Collaborated Reasoning (V-HOI MLCR) framework is shown in Figure \ref{fig:overview}. Given the video sequences, we use current a state-of-the-art V-HOI detection model to get initial prediction triplets for each frame. After that, we transfer the triplets to text format and feed them to our LLMs-based collaboration system. Our LLMs-based collaboration system involves two stages. 1) {\bfseries Cross-Agents Reasoning.} We define several distinct agents in each LLM (ChatGPT, LLaMA2, PaLM2), which are responsible for checking the prediction rationality in different aspects, e.g., spatial rationality and temporal rationality, and give respective confidence scores. 2) {\bfseries Multi-LLMs Debate.} We aggregate the responses from different LLMs through a Multi-LLMs debate format to get the final predictions.

\section{Cross-Agents Reasoning}

Benefiting from the large corpus and ample computing resources, LLMs achieve outstanding understanding performance in most natural language processing (NLP) tasks via incredible reasoning ability. Inspired by it, we aim to excavate the common sense, spatial, and temporal reasoning abilities of LLMs to facilitate the V-HOI task. To this end, we utilize each LLM as a cross-aspect reasoning agent to get more accurate predictions of existing V-HOI detection models. We give an overview of the Cross-Agents Reasoning process in Figure \ref{fig:agents}, in which we design three reasoning agents, i.e., common sense, spatial, and temporal reasoning agents, to refine the predictions obtained from an existing V-HOI detection model. For each agent, we prompted it to evaluate current HOI predictions. Each prompt includes a task description (Instruction) but with different roles to focus on different aspects when answering the question, a few examples (Demonstrations) that users carefully design, and a test instance (Test).

\subsection{Common Sense Reasoning Agent}
Given a predicted HOI triplet $<$person, relation, object$>$, e.g. $<$person, sit on, chair$>$, we first check if the relationship between the person and object is rational in common sense. Specifically, we ask LLM to score each prediction triplet based on common sense rationality. We first define the task description (Instruction):
\begin{quote}
    \it 
    You are an agent to give scores for all input examples based on their common sense rationality. Each input example is in the format $<$person, relation, object$>$. Your task is to score each input example based on the rationality of the relation between the person and the object. The output scores are between 0 and 1. Given an input example, you output the score. Please think step by step and then give the answer.
\end{quote}

\noindent{\bfseries In-context Learning.} We provide the LLM with manually curated examples after the task description. Through examples, we convey the exact format of our expectation for common sense reasoning and provide details of the instance specification. Examples (Demonstrations) are shown as follows:
 
\begin{quote}
    \it 
Input:$<$person,sit on,chair$>$ Output: 1.0 \\
Input:$<$person,sit on,table$>$ Output: 0.6\\
Input:$<$person,hug,table$>$ Output: 0.1\\
Input:$<$person,ride,elephant$>$ Output: 0.7\\
Input:$<$person,ride,bicycle$>$ Output: 1.0\\
\end{quote}

After the prompt and the examples, we ask the LLM to perform reasoning by given test instances (Test):

\begin{quote}
    \it 
Input: $<$person,hug,person$>$ Output: 1.0\\
Input: $<$person,hold,bicycle$>$ Output: 0.5\\
\end{quote}

The LLM is supposed to output the scores $\mathbf{s}_{cs}$ for each example triplet based on to what extent the example is aligned with human common sense.

\subsection{Spatial Reasoning Agent}
Besides the basic common sense reasoning ability of LLMs, enabling LLMs to understand the spatial correlations between the subjects and objects is also vital. As the input to LLM is only text, to incorporate the spatial information into the input, we involve another attribute of the subject and object when asking the LLM: The position (bounding boxes) of the subject and object, which explicitly gives the spatial locations at a video frame.

For the spatial agent, we check the triplet if it contains a relationship that is tidily related to the spatial locations of the subject and object. For instance, the action "ride" in triplet $<$person, ride, bike$>$ highly relies on the spatial location of the person and bike, as the person typically needs to be "above" or on the top of the bike in image space. On the contrary, the action "look at" in triplet $<$person, look at, person$>$ is not spatial aware given an image, as there is no certain location restriction of the action "look", i.e., a person can look a person from the right, left and anywhere.

To this end, there are two stages of our spatial reasoning assistant: 1) we ask LLMs if the predicted action in the triplet is spatial-aware. 2) we ask the assistant to give scores for example whether the predicted action is rational, based on the given subject and object locations. The task description (Instruction), examples (Demonstrations), and test instances (Test) are illustrated in Figure \ref{fig:agents} (b). We will get the evaluated scores $\mathbf{s}_{spatial}$ for each interaction prediction.

\subsection{Temporal Reasoning Agent}
We design a temporal reasoning agent to check the temporal rationality of a predicted triplet. Unlike the above-mentioned two agents which are applied to check all the testing examples, the temporal reasoning assistant is only applied to those that are changing between two frames. 

For example, if a triplet is $<$human, ride, bike$>$ at  $i-th$ frame, but changed to $<$human, carry, bike$>$ at the $(i+1)-th$ frame, then we will let temporal reasoning assistant to check if this change is reasonable. Specifically, we give the task description (Instruction), examples (Demonstrations), and test instances (Test) are illustrated in Figure \ref{fig:agents} (c), in which we could the evaluated scores $\mathbf{s}_{temporal}$. Finally, at each frame, we integrate all these scores (common sense score $\mathbf{s}_{cs}$, spatial score $\mathbf{s}_{spatial}$, and temporal score $\mathbf{s}_{temporal}$) with the initial model prediction $\mathbf{s}^{inter}$  from all three agents as follows:
\begin{equation}
    \begin{aligned} 
        \mathbf{s}_{final} =& ((\mathbf{s}^{inter} + \lambda_{cs} * sig(\mathbf{s}_{cs})) + \lambda_{s} * sig(\mathbf{s}_{spatial}))\\
        &+ \lambda_{t} * sig(\mathbf{s}_{temporal}).
    \end{aligned}
\end{equation}
where $sig(.)$ denotes the sigmoid function.

\section{Multi-LLMs Debate}
In our method, instead of only using one LLM to improve the prediction accuracy, we use several different LLMs to fully leverage the different knowledge. However, reasoning from the LLM sometimes gets incorrect answers, and some recent studies \cite{liang2023encouraging, gou2023critic} also prove that LLMs struggle to self-correct their responses without external feedback. To this end, we incorporate a debate scheme to integrate the responses from different LLMs. This debate process is illustrated in Algorithm \ref{alg:debate}. Generally, the framework is composed of two components, which are elaborated as follows:

\paragraph{LLMs as Debaters.}

Once we have received the initial responses, we designate each LLM to act as a participant in the debate. We have a total of $N$ participants, labeled as debaters, denoted by $D = \{D_i\}_{i=1}^N$ within our system. During the debate procedure, we prompt the first debater to provide an answer to the question. Then, the subsequent debaters, referred to as $D_i$, take their turns sequentially. Each one builds upon the debate history provided by the previous debater, denoted as $H$, so that the argument from debater $D_i$ given $H$ is represented as $D_i(H) = h$. An example of what a debater's prompt might look like:

\begin{quote}
    \it
     You are a debater among a panel of agents, each of whom will give their responses to the posed question in a debate setting. You do not need to fully agree with each other's perspectives, as our objective is to discuss and find the most reasonable answer. Please share your opinions in brief.
\end{quote}

\paragraph{LLM as a Judge.}

In addition, it is essential to designate an arbiter, referred to as $J$, to administer and supervise the debate sequence. In alignment with the framework suggested by \cite{liang2023encouraging},  we characterize the role of the judge $J$ to deduce the conclusive answer from the aggregate debate history, formalized as $J(H) = a$. A sample prompt for the judge is delineated as:

\begin{quote}
    \it
    You are a moderator. There will be three debaters involved in discussing a question. They will present their answers and discuss their perspectives on the correct answer. At the end of the debate, you will be responsible for deciding which answer is the most reasonable one based on the debate content. 
\end{quote}

As there are multiple LLMs involved but we only need one judge, we choose the generally most powerful LLM, ChatGPT, to play the role of judge.

\begin{algorithm}[t]
    \caption{Multi-LLMs Debate}\label{alg:debate}
    \begin{algorithmic}[1]
        \Require The question $q$ and number of debaters $N$
        \Ensure Final answer $a$
        
        \Procedure{MLD}{$q$, $N$}
            \State $J$ \Comment{Initialize the judge}
            \State $D \gets [D_1,\cdots,D_N]$ \Comment{Initialize debaters}
            \State $H \gets [q]$   \Comment{Initialize debate history}
            \For{each $D_i$ in $D$}
            \State $h \gets D_i(q)$  
            \Comment{Generate the initial answer}
            \State $H \gets H + [h]$ 
            \Comment{Append $h$ to $H$}
                \For{each $D_j$ in ${D \setminus D_i}$}
                    \State $h \gets D_j(H)$  \Comment{Generate argument}
                    \State $H \gets H + [h]$ \Comment{Append $h$ to $H$}
                \EndFor
            \EndFor
            \State $a \gets J(H)$         \Comment{Extract the final answer}
            \State \Return $a$
        \EndProcedure%
    
    \end{algorithmic}
\end{algorithm}

\section{Auxiliary Training with CLIP}
As we mentioned before, the existing V-HOI models have very limited reasoning ability due to the biased knowledge of specific datasets. Motivated by this, we plan to improve the internal reasoning capabilities of the prediction model itself further to ensure more seamless integration with LLMs.

\begin{figure}[ht]
    \centering
    \includegraphics[width=\linewidth]{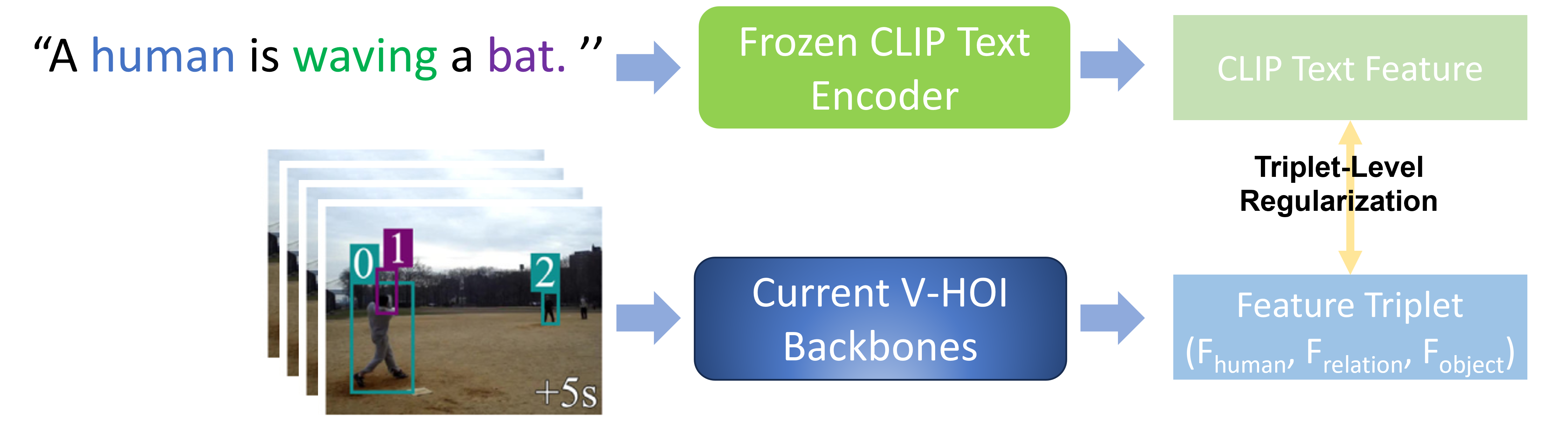}
    \caption{The auxiliary training strategy using CLIP feature for regularization.}
    \label{fig:cs}
\end{figure}

To this end, we involve an auxiliary training strategy as shown in Figure \ref{fig:cs} to enhance the existing V-HOI models using visual-linguistic knowledge captured by CLIP \cite{radford2021learning}. In particular, for every ground truth relationship triplet, we can produce a CLIP text embedding from its text format, and apply regularization to the corresponding triplet features $\vf_{human}, \vf_{inter}, \vf_{obj}$ at the end of feature extraction layer of the current V-HOI models. The CLIP text embeddings are offline extracted by the template ''A scene of a [person] [predicated interaction] a/an [object]'' for each GT relation. Subsequently, regularization aims to reduce the distance between the text embeddings $\ve_{ij}^\text{text}$ and the integrated triplet features $\vf_{ij}^\text{model}$:
\begin{equation}
    L_\text{tri-emb} = \sum_{i=1}^K \sum_{j=1,j\neq i}^K \rho(\vf_{ij}^\text{model}, \ve_{ij}^\text{text}) \cdot \Ibb_{[\ve_{ij}^\text{text}~\text{is from GT triplet}]} \label{eq:triplet_embedding_loss}
\end{equation}
where $\vf_{ij}^\text{model} = \mathtt{MLP}(\mathtt{cat}(\vf_{human}, \vf_{inter}, \vf_{obj}))$ is the fused embedding of the concatenated feature embeddings of detected human, interaction, and object.
$\rho(\cdot, \cdot)$ is a distance metric, we can apply $\ell_1$ norm or negative cosine distance.
$\Ibb_{[\cdot]}$ is an indicator function that equals to $1$ when the argument is true, and $0$ otherwise.
Thus, $L_\text{tri-emb}$ only regularizes the features whose triplets have ground-truth relations. 

We denote the original model loss as $L_\text{model}$, and the total loss $L_\text{total}$ is defined as:
\begin{equation}
    L_\text{total} = L_\text{model} + \lambda_{CLIP}L_\text{tri-emb} \label{eq:total_loss}
\end{equation}
where we set $\lambda_{CLIP}$ to balance the total loss because using the overly strong supervision of CLIP may disturb the original training target.

\section{Experiments}
\subsection{Datasets \& Metrics}
We evaluate our framework on two widely used V-HOI datasets: the Action Genome (AG)\cite{ji2020action} dataset for \textit{single-person} scenarios and the VidHOI \cite{chiou2021st} dataset for \textit{multi-person} scenarios. 

\textbf{Action Genome (AG)}\cite{ji2020action} dataset, a significant and large video dataset, provides detailed HOI annotations at the frame level. It contains annotations for 35 object classes, about 1.7M instances of 25 predicate classes, and 157 HOI categories, spanning 234K frames. These 25 predicates encompass three distinct types: attention predicates indicating a person's focus on a certain object, spatial predicates, and contact predicates illustrating different manners of object interaction like geometric position in space, contact, and non-contact relationships.

\textbf{VidHOI}\cite{chiou2021st} dataset emerges as the largest compilation of video data featuring comprehensive HOI annotations so far, which aggregates videos depicting unscripted human activities in dynamic, unstructured environments, sourced from social media—effectively portraying genuine real-world scenarios. This dataset encompasses 7122 videos, 78 object categories, and 50 relation classes including action and spatial relations. Note that half of the relation classes pertain to temporal relationships. It supplies 755K HOI instances and the most annotated keyframes (7.3M in total), and also defines the most HOI categories (557 in total) in the existing video datasets.

We report widely-used \textbf{Recall@K}\cite{ji2020action} (K = [10, 20, 50]) metric in scene understanding to quantify the model performance, which reveals the proportion of ground-truth occurrences found within the top-K confident predictions. Following \cite{cong2021spatial, ni2023human}, the setting of \textbf{\textit{Semi Constraint}} is used in our experiments which refers to the fact that all predictions with a confidence level higher than a predefined threshold are considered to be positive predictions.

\subsection{Implementation Details \& Baselines}
In our framework, we use the ChatGPT-4.0 model, the LLaMA2 model with 70B parameters, and the PaLM2 model with Bison size. Since our goal is to explore the accurate detection of relation/interaction with the aid of LLMs, in the experiments of this paper, we primarily focus on the \textbf{PREDCLS} setting which is that given ground truth labels and bounding boxes of humans and objects, predict relation/interaction labels of human-object pairs. To retrain the current V-HOI detection models with the auxiliary CLIP training strategy, we conduct all experiments on an RTX 3090 GPU. We test our method on the state-of-the-art V-HOI model STTranGaze \cite{ni2023human} and also test it on STTran \cite{cong2021spatial}. 

Since asking LLMs for each triplet of every frame of each video would result in more than ten million times of API calls over the entire dataset when it comes to the spatial and temporal agents and the debate scheme, and also the fact that the differences between neighboring frames are very small, for these reasons, we only ask LLMs for key frames that sampled from a fixed interval and then spread the results from LLMs to the non-key frames.

As for hyper-parameters, we set the CLIP-Loss weight $\lambda_{CLIP}$ to 0.05 on the VidHOI dataset and 1.5 on the AG dataset, and the weights for Common Sense, Spatial, and Temporal Scores are 0.05, 1.7, and 1.7 respectively on VidHOI dataset. For the AG dataset, we only apply Common Sense scores and the debate scheme for correcting models' predictions, both with a weight of 0.2 because the AG dataset doesn't have consistent identifiers to track the triplets. The threshold of confidence in the predicates/relations is set to 0.3 in the \textit{Semi Constraint} setting. All other hyper-parameters will be kept in line with the setting of STTranGaze \cite{ni2023human} and STTran \cite{cong2021spatial}.

\begin{table}[h]
\centering
\normalsize
\caption{Main Results on the VidHOI \& AG datasets}
\resizebox{\linewidth}{!}{
\begin{tabular}{c|c|ccc}
\hline
\multirow{2}{*}{Dataset} & \multirow{2}{*}{Method}                                                             & \multicolumn{3}{c}{Recall}                                                  \\
                         &                          & \multicolumn{1}{c}{@10} & \multicolumn{1}{c}{@20} & \multicolumn{1}{c}{@50} \\
                         \hline
\multirow{4}{*}{AG} 
& STTran\cite{cong2021spatial}     & 73.20    & 83.10    & 84.00    \\
                         & STTran+MLCR                         & \textbf{79.75}         & \textbf{93.25}          & \textbf{95.65}                        \\
                         & STTranGaze \cite{ni2023human}         &  75.40                       &   83.70         &   84.30                      \\
                         & STTranGaze+MLCR                 &  \textbf{79.61}     &  \textbf{93.69}   & \textbf{95.66}    \\
\hline
\multirow{4}{*}{VidHOI}  & STTran\cite{cong2021spatial}     & 65.77    &  69.61   &  70.71   \\
                         & STTran+MLCR     & \textbf{69.68}        &  \textbf{74.47}                       &  \textbf{76.22}                       \\
                         & STTranGaze \cite{ni2023human}         &  67.03    &  70.94  & 72.23                      \\
                         & STTranGaze+MLCR                               &  \textbf{70.47}                       &    \textbf{75.42}   &  \textbf{77.29}   \\
                         \hline
\end{tabular}}
\label{tab:compare}
\end{table}

\subsection{Experimental Results}

We report the main results on the AG dataset and the VidHOI dataset in Table \ref{tab:compare}. As shown, compared to state-of-the-art methods that we use as the baseline methods, the equipment of our MLCR method achieves superior performance, demonstrating the effectiveness of our framework. It is worth noting that even if only the common sense agent and debate scheme are applied, our V-HOI MLCR framework still obtains a huge performance gain compared to the two baseline models on the AG dataset under the metric R@50, respectively 13.87\%  for STTran with MLCR and 13.47\% for STTranGaze with MLCR. Meanwhile, the MLCR framework still shows promising performance when faced with the more complex multi-person scenario VidHOI dataset, the largest boosts continue to be seen in the R@50 metric, at 7.79\% for STTran with MLCR and 7.01\% for STTranGaze with MLCR, respectively. On the remaining metrics, the MLCR shows different levels of enhancement to the baseline models, which proves that the common sense, spatial-temporal corrections made by the MLCR to the predictions made by the baseline model using its powerful inference capabilities are reasonable and effective.

\subsection{Ablation Studies}

In this section, we analyze the effectiveness of the components of our framework in Table \ref{tab:ab_all}. All the ablation studies are conducted on the more complicated VidHOI dataset.

We first evaluate the proposed auxiliary CLIP training strategy with results shown in Table \ref{tab:ab_all}. We could observe that CLIP as a powerful internal supervisor can help semantic alignment and enhance the baseline V-HOI models’ reasoning and discriminative ability in handling ambiguous semantic relations, even though with a minor weight ($\lambda_{CLIP} = 0.05$) of CLIP Loss. Furthermore, table \ref{tab:ab_all} also shows the effectiveness of the common sense agent, spatial agent, and temporal agent. We could observe that as the three agent modules are added in turn, the performance of the baseline model progressively increases and is reflected in all evaluation metrics. Ultimately, the evaluation of the multi-LLMs debate scheme is still shown in Table \ref{tab:ab_all}. It could be observed that the debate mechanism can continue to robustly improve the performance of the MLCR framework, suggesting that this scheme makes the LLMs' answers more reasonable and their reasoning better. 

As we expected, different components of our framework all facilitate performance improvement. This further illustrates that the MLCR architecture has more outstanding common sense and spatio-temporal reasoning capabilities to effectively correct for biases and erroneous predictions of baseline models that have been trained on specific datasets.

\begin{table*}[ht]
\centering
\normalsize
\caption{Ablation Study on the VidHOI dataset over different components of our proposed framework.} 
\resizebox{\linewidth}{!}{
\begin{tabular}{c|c|ccc|c|ccc}
\hline
\multirow{2}{*}{Baseline} & \multirow{2}{*}{CLIP-Training} & \multicolumn{3}{c|}{Agents} & \multirow{2}{*}{Debate}    & \multicolumn{3}{c}{Recall} \\
                          & & Common Sense & Spatial & Temporal &                          & \multicolumn{1}{c}{@10} & \multicolumn{1}{c}{@20} & \multicolumn{1}{c}{@50} \\
\hline
\multirow{6}{*}{STTran \cite{cong2021spatial}}  &  &             &         &          &       & \textcolor{blue}{65.77}    &  \textcolor{blue}{69.61}   & \textcolor{blue}{70.71}    \\
& \checkmark &             &         &          &                              & 66.11    &  69.79   & 71.17    \\
& \checkmark & \checkmark            &         &          &                              & 67.25    &  71.39   & 72.79    \\
                         & \checkmark & \checkmark            & \checkmark       &          &                                             &  68.15                       & 72.53      &   74.02                      \\
                          & \checkmark& \checkmark            & \checkmark       & \checkmark        &                             &    68.63        &    73.10                     &   74.62                      \\
                          & \checkmark& \checkmark            & \checkmark       & \checkmark        & \checkmark                              &    \textbf{69.68}                  &   \textbf{74.47}     & \textbf{76.22}      \\
\hline
\multirow{6}{*}{STTranGaze\cite{ni2023human}}   &  &             &         &          &   & \textcolor{blue}{67.03}    &  \textcolor{blue}{70.94}   & \textcolor{blue}{72.23}    \\
& \checkmark &             &         &          &                              & 67.13    &  71.04   & 72.31    \\
& \checkmark& \checkmark            &         &          &                             &  68.26   & 72.50    & 73.69    \\
                          & \checkmark& \checkmark            & \checkmark       &          &                               &    69.08                     &    73.58                     &    75.17                     \\
                          & \checkmark& \checkmark            & \checkmark       & \checkmark        &                                            &    69.48                     &      74.05     &   75.66                      \\
                          & \checkmark& \checkmark            & \checkmark       & \checkmark        & \checkmark                             &  \textbf{70.47}                       &   \textbf{75.42}     &  \textbf{77.29}      \\ \hline 
\end{tabular}}
\label{tab:ab_all}
\end{table*}

\subsection{Visualization}
\begin{figure*}
    \centering
    \includegraphics[width=\textwidth]{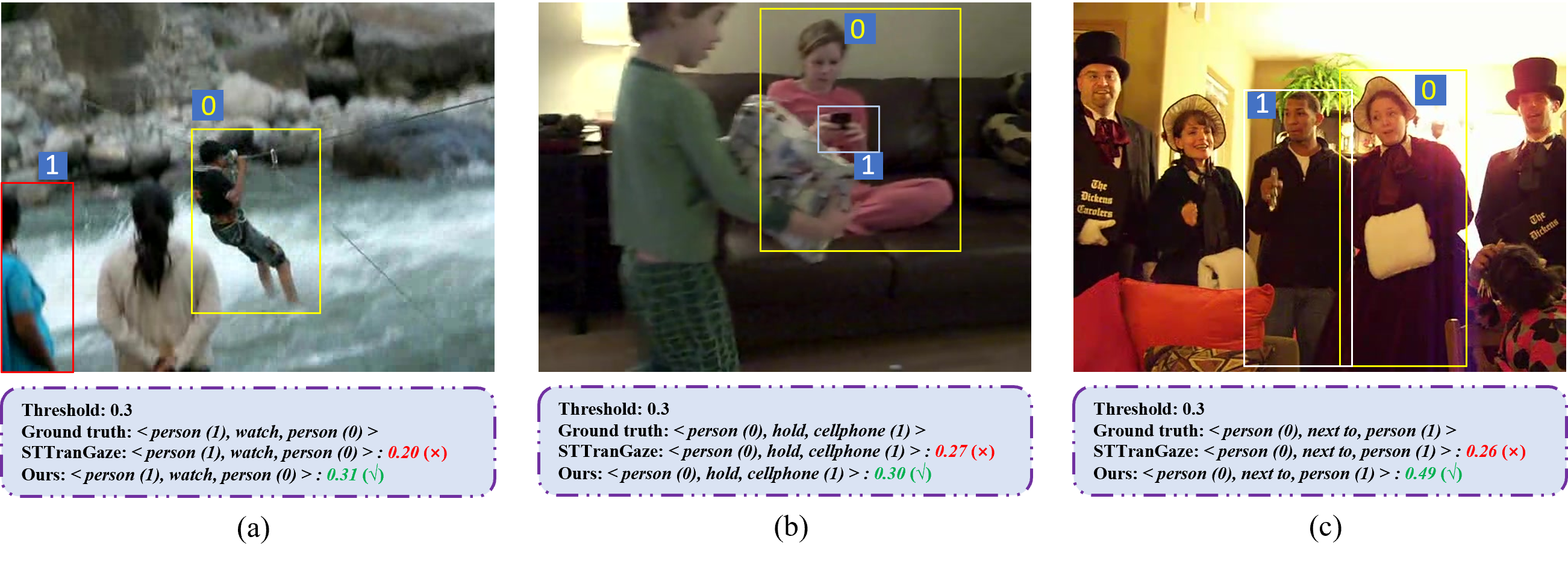}
    \caption {Some visual results of V-HOI MLCR. Our framework uses the scores of LLMs to raise the score of potential correct triplet and above the preset threshold so that they are judged as correct predictions that match the label.}
    \label{fig:vis}
\end{figure*}

In this section, we show some visual samples of our MLCR results to intuitively represent the superiority of multiple LLMs collaborated reasoning. As shown in Figure \ref{fig:vis}, our framework succeeds in correctly predicting relationships containing spatio-temporal information (e.g "watch", "hold", "next to") through the inference of LLMs, which pulls the score of the triplet up making it higher than the threshold, thus allowing the correct triplet to be accurately admitted. 

\section{Conclusion}
In this paper, we have proposed a novel framework V-HOI MLCR for Human-Centered Dynamic Scene Understanding. Specifically, we collaborate the existing V-HOI models with LLMs as external experts to facilitate their reasoning abilities and propose an auxiliary training scheme with CLIP to enhance baseline models. Our proposed framework is simple yet effective and can be used conveniently in a plug-and-play manner. Our framework achieves superior performance on the Action Genome dataset and the VidHOI dataset.

\bibliographystyle{IEEEtran}
\bibliography{egbib}

\end{document}